\documentclass[journal]{IEEEtran}

\usepackage{cite}
\usepackage{amsmath,amssymb,amsfonts}
\usepackage{algorithmic}
\usepackage{graphicx}
\usepackage{mathtools}
\usepackage{textcomp}
\usepackage{xcolor}
\usepackage{tabularx}
\usepackage{multirow}
\usepackage{adjustbox}
\usepackage{array}
\usepackage{makecell}
\usepackage{caption}

\usepackage{booktabs}
\usepackage{hyperref}
\usepackage{longtable}
\usepackage{graphicx} 
\usepackage{pifont}   
\usepackage{array}    

\ifCLASSINFOpdf
\else
\fi

\begin{document}
%
\title{MetaGreen: Meta-Learning Inspired Transformer Selection for Green Semantic Communication}
%
%
%


\author{\IEEEauthorblockN{Shubhabrata Mukherjee, Cory Beard, and Sejun Song}
\IEEEauthorblockA{\textit{School of Science and Engineering, University of Missouri-Kansas City, Kansas City, MO, USA}
 \\
smpw5,beardc,songsej@umsystem.edu}
}

\maketitle

\begin{abstract}
Semantic Communication can transform the way we transmit information, prioritizing meaningful and effective content over individual symbols or bits. This evolution promises significant benefits, including reduced latency, lower bandwidth usage, and higher throughput compared to traditional communication. However, the development of Semantic Communication faces a crucial challenge: the need for universal metrics to benchmark the joint effects of semantic information loss and energy consumption. This research introduces an innovative solution: the ``Energy-Optimized Semantic Loss'' (EOSL) function, a novel multi-objective loss function that effectively balances semantic information loss and energy consumption. Through comprehensive experiments on transformer models, including energy benchmarking, we demonstrate the remarkable effectiveness of EOSL-based model selection. We have established that EOSL-based transformer model selection achieves up to 83\% better similarity-to-power ratio (SPR) compared to BLEU score-based selection and 67\% better SPR compared to solely lowest power usage-based selection. Furthermore, we extend the applicability of EOSL to diverse and varying contexts, inspired by the principles of Meta-Learning. By cumulatively applying EOSL, we enable the model selection system to adapt to this change, leveraging historical EOSL values to guide the learning process. This work lays the foundation for energy-efficient model selection and the development of green semantic communication.
\end{abstract}

\begin{IEEEkeywords}
Green Semantic Communication, Transformer, Meta-Learning, Energy Optimized Loss Function, Large language Models
\end{IEEEkeywords}

%
\IEEEpeerreviewmaketitle
\section{Introduction}
\label{sec:intro}
\begin{figure*}[hbt!]
\centerline{\includegraphics[width=\textwidth]{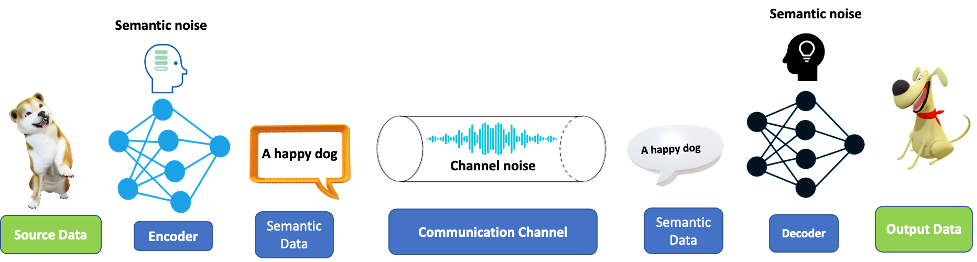}}
\caption{The basic blocks of a semantic communication}
\label{fig:SemComm_concept}
\end{figure*}

What is ``semantic"? The word ``semantic" comes from the ancient Greek adjective `semantikos', which means ``relating to signs" or ``significant". In modern communication, semantics goes beyond the dictionary definition of words, delving into how we understand individual words and phrases, the influence of context on their meaning, and the shared knowledge we rely on to decipher the message. Semantic Communication (SemCom) is a novel communication model that focuses on transmitting only semantically-significant information through a communication channel (Fig.~\ref{fig:SemComm_concept})~\cite{getu2023making}. The Shannon-Weaver model~\cite{shannon1948mathematical} identified three levels of communication paradigm:
\begin{itemize}
    \item \textbf{Technical Level}: This level addresses accuracy issues in message transmission, such as poor phone connections, typos in emails, or static in radio signals. Our current communication model falls within this realm.
    \item \textbf{Semantic Level}: This level delves into the meaning of messages, focusing on whether the sender and receiver share a mutual understanding of words and symbols. Challenges like misunderstandings of slang, cultural references, or jargon arise at this level. Semantic communication explores this facet of communication.
    \item \textbf{Effectiveness Level}: This level evaluates whether messages achieve the sender's objectives. Even when messages are clear and understood accurately, the receiver's response may not align with the sender's intent. Factors such as failed persuasion attempts, unclear instructions, or emotional responses can disrupt communication. Goal-oriented or intent-based communication can elucidate this level.
\end{itemize}

So far researchers have primarily focused on optimizing the technical level of communication. However, the rapid expansion of intricate AI-generated content intensifies the challenge of information overload within communication networks. In this context, Shannon's channel capacity theorem, expressed by the formula \( C = B \log_2(1 + \text{SNR}) \), highlights the finite capacity ($C$) of communication channels for an SNR level that can be achieved for a fixed bandwidth ($B$). Bandwidth limitations present a significant hurdle, potentially impeding the flow of information. The objective of SemCom is to convey the intended message meaning efficiently, either through text or other modal attributes. This approach aims to minimize power usage, bandwidth consumption, and transmission delays. By eliminating extraneous information that does not contribute to the message's meaning, SemCom enhances information transmission and communication performance. Traditional communication methods, such as image transmission, prioritize data completeness over efficiency. Semantic communication offers a solution by transmitting only the essential data essence, often through textual descriptions. By streamlining communication through prioritizing meaning over raw data, semantic goal-oriented communication not only reduces bandwidth usage and energy consumption, addressing information overload and bandwidth limitations but also goes beyond mere data transmission. It ensures clarity of understanding and attainment of desired outcomes – aspects typically associated with the semantic and effectiveness levels of the Shannon communication model.

The Generative AI revolution, fueled by ubiquitous attention-based architectures like transformers and large language models (LLMs)~\cite{vaswani2017attention}, opens the door for the development of powerful semantic encoders and decoders. Several image captioning models based on transformers, such as Bootstrapping Language-Image Pre-training (BLIP)~\cite{li2023blip2} and Vision Transformer (ViT)~\cite{dosovitskiy2020image}, excel at transforming images into text. These models could potentially be utilized as the semantic encoder in a semantic communication system. Conversely, text-to-image generation models with transformer architectures, like Stable Diffusion~\cite{rombach2022high} and SDXL-Lightning~\cite{lin2024sdxl}, could function as the decoder, reconstructing an image based on the encoded semantic meaning. Transformers are increasingly becoming the go-to solution for diverse problems in AI. However, their reliance on attention weight mechanisms makes them computationally expensive, especially large foundation LLMs. Trained on massive amounts of data, these LLMs rank among the most power-hungry deep learning models ever built. While advancements in hardware accelerate training times, the growing complexity of models still translates to significant energy consumption. This necessitates the development of more efficient transformer architectures that minimize environmental impact and enable deployment in resource-constrained environments. LLMs and transformer models are well-known for their significant energy consumption~\cite{mcdonald2022great}. To address this issue, researchers are actively investigating various techniques to enhance their efficiency. These include approaches such as one bit LLMs, weight quantization of diffusion model, quantized federated learning etc.~\cite{ma2024era,sui2024bitsfusion, kim2023green, zou2023wireless}. However, current methods often rely on indirect metrics such as FLOPs (floating-point operations) or estimated training energy~\cite{zhou2020hulk}. In our ongoing research, we have taken a different approach by directly measuring CPU, GPU, and system utilization during inference. This enables us to provide a more precise assessment of energy consumption, facilitating better comparisons between different models. Our findings reveal that a typical text-to-image conversion using a diffusion model consumes approximately 4 kJ of energy. This translates to an estimated emission of 0.5 grams of CO$_2$, which is equivalent to burning 0.23 grams of coal or heating 10 ml of water from room temperature to boiling~\cite{EPA_GHG_Calculator}.

While minimizing energy consumption is crucial, another key challenge lies in ensuring the fidelity of the information being transmitted. To construct a reliable and efficient semantic communication system, it's crucial to understand the encoding and decoding capabilities of the transformer models used within it. Researchers have pursued various methods to quantify the loss of semantic information during the end-to-end encoding and decoding of semantic messages. In communication systems, semantic transformation loss refers to the degradation or alteration of the transmitted data's meaning or information content. This loss can occur due to differences in data interpretation between the sender and receiver, or due to errors that arise during transmission. The consequences of semantic transformation loss may include misunderstandings, misinterpretations, or incomplete information, all of which can lead to inefficiencies or errors in communication. Researchers are tackling this challenge by employing diverse metrics such as Structural Similarity Index Measure (SSIM), Word Error Rate (WER), Peak Signal-to-Noise ratio (PSNR), and Kullback–Leibler divergence (KID) \cite{sun2013improved}. However, they did not consider the correlation between energy consumption and semantic loss. To develop an innovative and energy-efficient ``Green Semantic Communication System," it's essential to find a balance between semantic capability and energy efficiency. A holistic framework for deep learning based SemCom, including performance metrics and suitable AI architecture are crucial~\cite{luo2022semantic}. Our current research presents a multi-objective loss metric function named Energy-Optimized Semantic Loss (EOSL) and thus offers a more robust and comprehensive SemCom system model. Our study reveals that informed model selection notably improves semantic efficiency while minimizing resource requirements. The EOSL metric could be integrated into LLM comparison frameworks like Google's recent LLM Comparator~\cite{kahng2024llm} to create a more holistic evaluation that considers both performance and energy efficiency. This would be beneficial not just for semantic communication systems, but for any application that utilizes LLMs where both fidelity and resource consumption are important factors.

The unique contributions of this paper include:

\begin{itemize}

    \item A comprehensive comparison study of existing performance metrics, summarizing their capabilities and features, and demonstrating the superiority of EOSL in balancing semantic information loss and energy consumption.
    
    \item Introduction of Energy Optimized Semantic Loss (EOSL), a novel multi-objective loss function guiding transformer model selection for improved semantic efficiency without excessive energy.
    
    \item Extensive benchmarking of transformer models' resource utilization, including CPU and GPU energy usage, and validating EOSL's efficiency in selecting optimal models for semantic encoding and decoding through simulation studies.

    \item Successful application of Meta-Learning principles to extend EOSL's applicability to diverse contexts without additional backpropagation, enhancing its adaptability and sustainability.
\end{itemize}

In this paper, Section \ref{sec:review} presents a comprehensive review on the trend of increasing machine learning task complexity, along with the model size, computation complexity and energy consumption. It also discusses the recent work on energy efficient semantic communication. Section \ref{sec:limitation} discusses the limitations of existing metrics used to evaluate models' semantic efficiency and energy consumption. In Section \ref{sec:energy_optimized_semantic_loss} we introduce the Energy-Optimized Semantic Loss (EOSL) function; \ref{sec:cumulative_learning} discusses generalization capability of EOSL; and \ref{sec:semantic_encoder_decoder}  discusses building semantic communication encoders and decoders using transformers.  Section \ref{sec:results} provides our results and discussion, and finally Section \ref{sec:conclusion} concludes the paper with insights and potential future research directions.

\section{Literature Review}
\label{sec:review}
\begin{figure}[hbt!]
\centering
\includegraphics[width=\columnwidth]{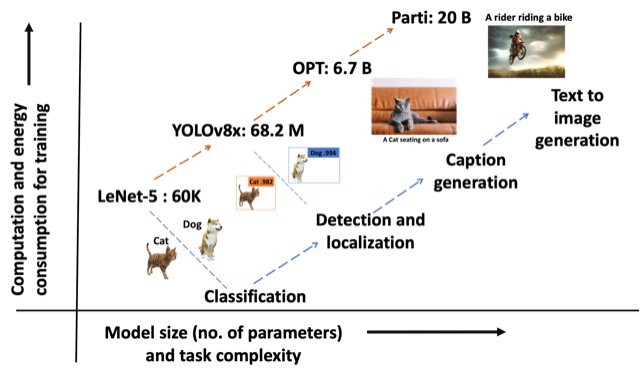} 
\caption{Evolution of complexity and training requirement of models}
\label{fig:SemComm_complexity}
\end{figure}

Deep learning models have grown significantly in size and computational requirements over time, driven by the need to perform more complex tasks that demand higher model complexity and larger data sets~\cite{hu2021model}. Early deep-learning models had only a few layers and limited parameters, primarily used for basic image and speech recognition. However, as the field has progressed, larger and more complex models have been developed to tackle more challenging problems, such as natural language processing, computer vision, and speech synthesis. Looking ahead to the near future, the trend of increasing model complexity and computational requirements are expected to continue. Fig.~\ref{fig:SemComm_complexity} shows LeNet~\cite{lecun1998gradient} using only 60k parameters for image classification, object detection using YOLOv8x~\cite{Jocher_YOLO_by_Ultralytics_2023} with 68 M parameters, OPT~\cite{zhang2022opt} for caption generation using 6.7 B, and Parti~\cite{yu2022scaling}, a text-to-image generation model by Google can scale up to 20 billion parameters. The well-known AlexNet architecture introduced in 2012 had around 60 million parameters, whereas modern state-of-the-art Large Language Models (LLM) like GPT-3 and EfficientNet can have billions of parameters. EfficientNet-B0, for instance, has only 5.3 million parameters but can achieve state-of-the-art accuracy on ImageNet with 6.4 times fewer FLOPs than the previous state-of-the-art model, while using 8.4 times less memory. While deep learning models have become faster to train with better hardware, their growing complexity still demands significant energy. This necessitates choosing efficient models that minimize environmental impact and enable deployment in resource-limited settings.
MobileNet exemplifies energy-efficient design for mobile and similar contexts. The original MobileNet model had only 4.2 million parameters and could be trained with as little as 500,000 images, much smaller than other state-of-the-art models for image recognition. In anexperiment,~\cite{garcia2019estimation} showed MobileNet as the most energy-efficient ConvNet choice under similar execution environments compared to Inception-V3 and DenseNet. EfficientNet is another energy-efficient deep-learning model that achieves state-of-the-art performance while using fewer parameters and less computation. It achieves this by using a novel compound scaling method that scales the model's depth, width, and resolution in a principled way. EfficientNet-B0 has only 5.3 million parameters and an energy efficiency of 4.6 billion operations per joule, which is much higher than other state-of-the-art models. Energy-efficient deep learning models like MobileNet and EfficientNet are ideal for resource-constrained environments as they achieve state-of-the-art performance with relatively low computational requirements and energy usage. Researchers have recently proposed various energy-efficient semantic communication architectures for aerial edge networks, including those utilizing rate splitting techniques and either energy-aware computation offloading to edge servers or energy-aware content caching techniques. However, while these studies focus on structural aspects, they fail to explicitly address the raw energy consumption and its balance with semantic efficiency.~\cite{10527365,yang2023energy,saadat2024energy}

\section{Limitations of Existing Metrics}
\label{sec:limitation}
Table~\ref{tab:metrics_techniques_comparison} provides a comprehensive comparison of existing metrics and techniques for evaluating and optimizing model performance, size, computation, and speed. While metrics like Inception Score, SSIM, or Bilingual Evaluation Understudy (BLEU) Score focus on a model's ability to perform its assigned task with high accuracy (e.g., text-to-text transformation), they do not explicitly account for the energy expenditure required to achieve that task. On the other hand, indirect metrics like GFLOP and Energy Delay Product measure computational or energy efficiency, but do not necessarily provide insights into a model's performance capabilities. Techniques like Knowledge Distillation, Quantization, and Pruning optimize model size, execution efficiency, and hardware utilization, enabling efficient deployment across various platforms; however, these techniques often require additional optimization and regularization through backpropagation, which can lead to increased energy consumption. In contrast, our proposed loss function, EOSL, uniquely addresses both performance similarity and energy efficiency without requiring additional optimization processes. This makes EOSL a superior and effective solution for evaluating and optimizing model performance.

\begin{table*}[ht]
\caption{Metrics and Technique Comparisons}
\centering

\begin{tabular}{|m{3.5cm}|m{6.5cm}|c|c|c|c|}
\hline    &                              & \textbf{Image} & \textbf{Text} & \textbf{Model} &  \\\textbf{Metric/Technique}                                                        & \centering\textbf{Capability/Feature}                                                                                                                              & \textbf{Semantic} & \textbf{Semantic} & \textbf{Energy} & \textbf{Additional} \\& & \textbf{Similarity}       & \textbf{Similarity}    & \textbf{Efficiency}    & \textbf{Optimization}             \\ \hline
\textbf{Energy Optimized Semantic Loss (ours)}                          & Expresses the joint effect of direct energy efficiency and semantic similarity                                                                            & \checkmark                                & \checkmark                                & \checkmark                               & \ding{55}                                           \\ \hline
Inception Score~\cite{salimans2016improved}                                                        & Measures the quality and diversity of the generated images                                                                                                   & \checkmark                                  & \ding{55}                                & \ding{55}                                & —                                           \\ \hline
Mean Squared Error, Root Mean Squared Error              & Measures pixel-wise difference between images                                                                                                             & \checkmark                                 & \ding{55}                                & \ding{55}                                & —                                           \\ \hline
Peak Signal-to-Noise Ratio                                   & Measures the ratio between the maximum possible signal and the background noise                                                                           & \checkmark                                 & \ding{55}                                & \ding{55}                                & —                                           \\ \hline
Structural Similarity Index Measure (SSIM)~\cite{wang2004image}                            & Considers perceived image quality factors like luminance, contrast, and structure                                                                         & \checkmark                                 & \ding{55}                                & \ding{55}                                & —                                           \\ \hline
Word Mover's Distance~\cite{kusner2015word}                                           & Measures semantic distance between text documents based on word embeddings                                                                                & \ding{55}                                  & \checkmark                               & \ding{55}                                & —                                           \\ \hline
Cosine Similarity                                                     & Measures similarity between text documents based on the cosine of the angle between their vector representations                                           & \checkmark                                 & \checkmark                               & \ding{55}                                & \ding{55}                                           \\ \hline
Jaccard Similarity~\cite{niwattanakul2013using}                                                    & Measures similarity between text documents based on the intersection over union of their word sets                                                        & \ding{55}                                  & \checkmark                               & \ding{55}                                & \ding{55}                                           \\ \hline
BLEU Score~\cite{papineni2002bleu}                                                            & Measures similarity between machine-translated text and a reference translation                                                                           & \ding{55}                                  & \checkmark                               & \ding{55}                                & —                                           \\ \hline
ROUGE Score~\cite{lin2004rouge}                                                           & ROUGE scores measure text similarity by evaluating overlap of n-grams with a reference text                                                               & \ding{55}                                  & \checkmark                               & \ding{55}                                & \ding{55}                                           \\ \hline
Embedding Similarity Metrics                                          & Measure similarity between pre-trained word or image embeddings                                                                                           & \ding{55}                                  & \checkmark                               & \ding{55}                                & —                                           \\ \hline
Earth Mover's Distance (EMD)~\cite{rubner1998metric}                                          & Similar to WMD, but uses transportation cost matrix to account for semantic relationships between words                                                   & \checkmark                                 & \ding{55}                                & \ding{55}                                & —                                           \\ \hline
Low-Rank Approximation Techniques~\cite{hu2021lora} & Low-Rank Approximation (LoRA) compresses high-dimensional data by capturing its essence in a lower-dimensional space                                      & \checkmark                                 & \checkmark                               & \ding{55}                                & \checkmark                                          \\ \hline
Floating-point Operations                                     & Counts the number of basic mathematical operations performed during training or inference. Lower FLOPs indicate potentially lower energy consumption          & \ding{55}                                  & \ding{55}                                & \checkmark                               & —                                           \\ \hline
Training Time                                                         & Measures the time it takes to train a model. Faster training times can translate to lower energy usage                                                   & \ding{55}                                  & \ding{55}                                & \checkmark                               & —                                           \\ \hline
Hardware Utilization                                                  & Tracks how efficiently the hardware (GPUs, TPUs) is utilized during training or inference                                                                & \ding{55}                                  & \ding{55}                                & \checkmark                               & —                                           \\ \hline
Energy Consumption (Watts)                                            & Directly measures the power consumption of the hardware running the model                                                                                  & \ding{55}                                  & \ding{55}                                & \checkmark                               & \checkmark                                          \\ \hline
Energy Delay Product (EDP)~\cite{laros2013energy}                                            & Combines energy consumption and latency (execution time). Lower EDP indicates a more energy-efficient model for a given task                             & \ding{55}                                  & \ding{55}                                & \checkmark                               & —                                           \\ \hline
Knowledge Distillation~\cite{hinton2015distilling}                                                & Trains a smaller student model by mimicking the predictions of a larger teacher model. This can achieve similar accuracy with lower energy requirements & \ding{55}                                  & \ding{55}                                & \checkmark                               & \checkmark                                          \\ \hline
Quantization~\cite{jacob2018quantization}                                                          & Reduces the precision of weights and activations in the model while minimizing accuracy loss                                                             & \ding{55}                                  & \ding{55}                                & \checkmark                               & \checkmark                                          \\ \hline
Pruning-Quantization-aware Training~\cite{hawks2021ps}                               & Combines parameter pruning and quantization techniques during training itself, leading to more efficient models from the start                              & \ding{55}                                  & \ding{55}                                & \checkmark                               & \checkmark                                          \\ \hline
\end{tabular}

\label{tab:metrics_techniques_comparison}
\end{table*}

\section{Energy Optimized Semantic Loss}
\label{sec:energy_optimized_semantic_loss}
Driven by the challenge of balancing semantic reliability and energy efficiency in semantic communication, we introduce the ``Energy-Optimized Semantic Loss" (EOSL) function. This multi-objective metric captures both semantic information loss and the energy requirements of the semantic communication process, facilitating informed model selection and resource optimization. Next, we outline the development process of EOSL.

First, we introduce a method for measuring semantic noise by quantifying semantic similarity. Let's denote the intended meaning of a message as $M_i$ and the perceived meaning as $M_p$. Then, the degree of semantic similarity in the message can be represented as:

\begin{equation}
S_{sm} = f(M_i, M_p)
\end{equation}

\noindent where $0 \leq S_{sm} \leq 1$. Here, $f()$ is a function measuring the similarity between the intended and perceived meanings. A smaller value of $f()$ indicates a greater amount of semantic noise in the message. Typically, $f()$ is computed using semantic similarity metrics, such as cosine similarity and structural similarity index measure (SSIM), comparing the original input message and the final output message.

Cosine similarity, denoted as $\cos({\bf A},{\bf B})$, between any two vectors is expressed as:

\begin{equation}
\cos ({\bf A},{\bf B}) = \frac{{\bf A} \cdot {\bf B}}{\|{\bf A}\| \|{\bf B}\|} = \frac{{\sum_{i=1}^{n}{{\bf A}_i{\bf B}_i}}}{{\sqrt{\sum_{i=1}^{n}{({\bf A}_i)^2}} \sqrt{\sum_{i=1}^{n}{({\bf B}_i)^2}}}}
\end{equation}

\noindent where $\textbf{A}$ and $\textbf{B}$ are the two image or word embedding vectors being compared. SSIM, on the other hand, compares two images based on structural similarity. Its formula is defined as:

\begin{equation}
SSIM(x,y) = \frac{(2\mu_x\mu_y + C_1) + (2 \sigma _{xy} + C_2)}{(\mu_x^2 + \mu_y^2+C_1) (\sigma_x^2 + \sigma_y^2+C_2)}
\end{equation}

\noindent Here, $\mu_x$ and $\mu_y$ represent the pixel sample means of $x$ and $y$ respectively, while $\sigma_x^2$ and $\sigma_y^2$ denote their variances. $\sigma_{xy}$ is the covariance of $x$ and $y$, and $C_1$ and $C_2$ are variables stabilizing the division. Consequently, the semantic noise can be expressed as:

\begin{equation}
N_{sm} = 1 - S_{sm} (M_i, M_p) \label{eq:Nsm}
\end{equation}

The specific form of the function $f()$ and the choice of semantic similarity metric may vary depending on the specific application and context in which semantic noise is being measured. Next, we consider the effects of communication channel noise on the accuracy of the received message. Given is a probability of bit error $p_b$ caused by random channel noise and the average $E_b/N_0$ in a particular environment. Also, there is a probability $p_{f}$ of being in a deep fade, in which case all bits are lost (coincidentally the probability $0.5$ of bits in error). Given is the following average probability of bit error $\bar{p_b}$.

\begin{equation}
    \bar{p_b}=0.5p_{f}+p_b(1-p_{f})
\end{equation}

Next, we create a channel loss component to compare traditional and semantic communications by assuming a channel loss $L_{ch}$ analogous to the block error rate in the presence of a channel coding that can correct up to $t$ errors. $L_{ch}$ can be expressed as:

\begin{equation}
L_{ch}=1-\sum_{i=0}^{t} {l\choose i} {\bar{p_b}}^i (1-{\bar{p_b}})^{l-i}
\end{equation}

\noindent where $l$ is the length of the packet in bits.

We can compute the communication energy required for transmission using both traditional and semantic communication methods, highlighting the significant reduction in communication energy achieved with SemCom. We represent the communication energy used in both traditional and semantic approaches using the term $C_e$. Finally, the semantic energy referred to as $M_e$, which can be defined as the energy required for encoding or decoding semantic messages from one form to another. Hence EOSL can be written as below:

\begin{equation}
    EOSL = \sum_{j=1}^{n} \left\{ \lambda_{sm}(N_{sm_j}) + \lambda_{lch}(L_{ch_j}) + \right.
\left. \lambda_{e_{c}}(C_{e_j}) + \lambda_{e_{s}}(M_{e_j}) \right\}
\label{eq:EOSL2}
\end{equation}

\noindent  weight multipliers, denoted as $ \lambda_{sm} $, $ \lambda_{lch} $, $ \lambda_{e_{c}} $, and $ \lambda_{e_{s}} $, to offer flexibility in adjusting the influence of the losses relative to other terms. 

In this context, $n$ represents the total number of re-transmissions until the requirement reduction in semantic noise, $N_{sm_j} \leq N_{sm_{\text{thresh}}}$, is satisfied. The process of encoding, transmitting encoded messages, and decoding iterates until the condition is met. Here, $N_{sm_{\text{thresh}}}$ signifies the predetermined threshold for semantic noise. It is important to clarify that in this context, the EOSL is not employed as a training loss or regularization term. Its primary purpose is the selection of the most effective transformer model, serving either as an encoder or decoder, based on criteria that combine both semantic similarity and energy consumption. Consequently, it does not introduce any additional constraints or optimizations into the training procedure of the individual transformer model. EOSL normalizes both energies by dividing them by their maximum energy values $E_{c,max}$, and $E_{s,max}$ respectively among all available encoder/decoder options. The goal of our system is to find the model with the smallest EOSL. The EOSL can be expressed as shown in Equation~(\ref{eq:EOSL_full}). All the components used to construct the EOSL are summarized in Table~\ref{table:eosl_terms}.

\vspace{-0.2cm}
\begin{equation}
\begin{aligned}
EOSL &= \sum_{j=1}^{n} \left\{ \lambda_{sm}\Bigl(1 - S_{sm_j}(M_i, M_p)\Bigr) +  \lambda_{lch}\Bigl(L_{ch_j}\Bigr) + \right. \\
& \left. \lambda_{e_c}\Bigl(\frac{E_{c_j}}{E_{c,\text{max}}}\Bigr) + \lambda_{e_s}\Bigl(\frac{E_{s_j}}{E_{s,\text{max}}}\Bigr) \right\}
\end{aligned}
\label{eq:EOSL_full}
\end{equation}

\begin{table*}[h!]
\centering
\begin{tabular}{|c|p{10cm}|}
\hline
\textbf{Term} & \textbf{Description} \\ \hline
\( n \) & Total number of retransmissions\\ \hline
\( \lambda_{sm} \) & Weight multiplier for the semantic noise component. \\ \hline
\( \lambda_{lch} \) & Weight multiplier for the channel loss component. \\ \hline
\( \lambda_{ec} \) & Weight multiplier for the communication energy component. \\ \hline
\( \lambda_{es} \) & Weight multiplier for the semantic energy component. \\ \hline
\( S_{sm_j}(M_i, M_p) \) & Semantic similarity score between the transmitted and the received message \( M_i \),\( M_p \). \\ \hline
\( L_{ch_j} \) & Communication Channel loss for the \( j \)-th transmission. \\ \hline
\( E_{c_j} \) & Communication energy used in the \( j \)-th transmission. \\ \hline
\( E_{c_{max}} \) & Maximum communication energy among all encoder/decoder options. \\ \hline
\( E_{s_j} \) & Semantic energy used in the \( j \)-th transmission. \\ \hline
\( E_{s_{max}} \) & Maximum semantic energy among all encoder/decoder options. \\ \hline
\end{tabular}
\caption{Summary of Terms Used in The EOSL Formula}
\label{table:eosl_terms}
\end{table*}

EOSL can now be used to compare SemCom encoders with each other and involve  communications. The Decoder incurs even higher energy usage, discussed later.

\begin{figure*}[hbt!]
\centerline{\includegraphics[width=\textwidth]{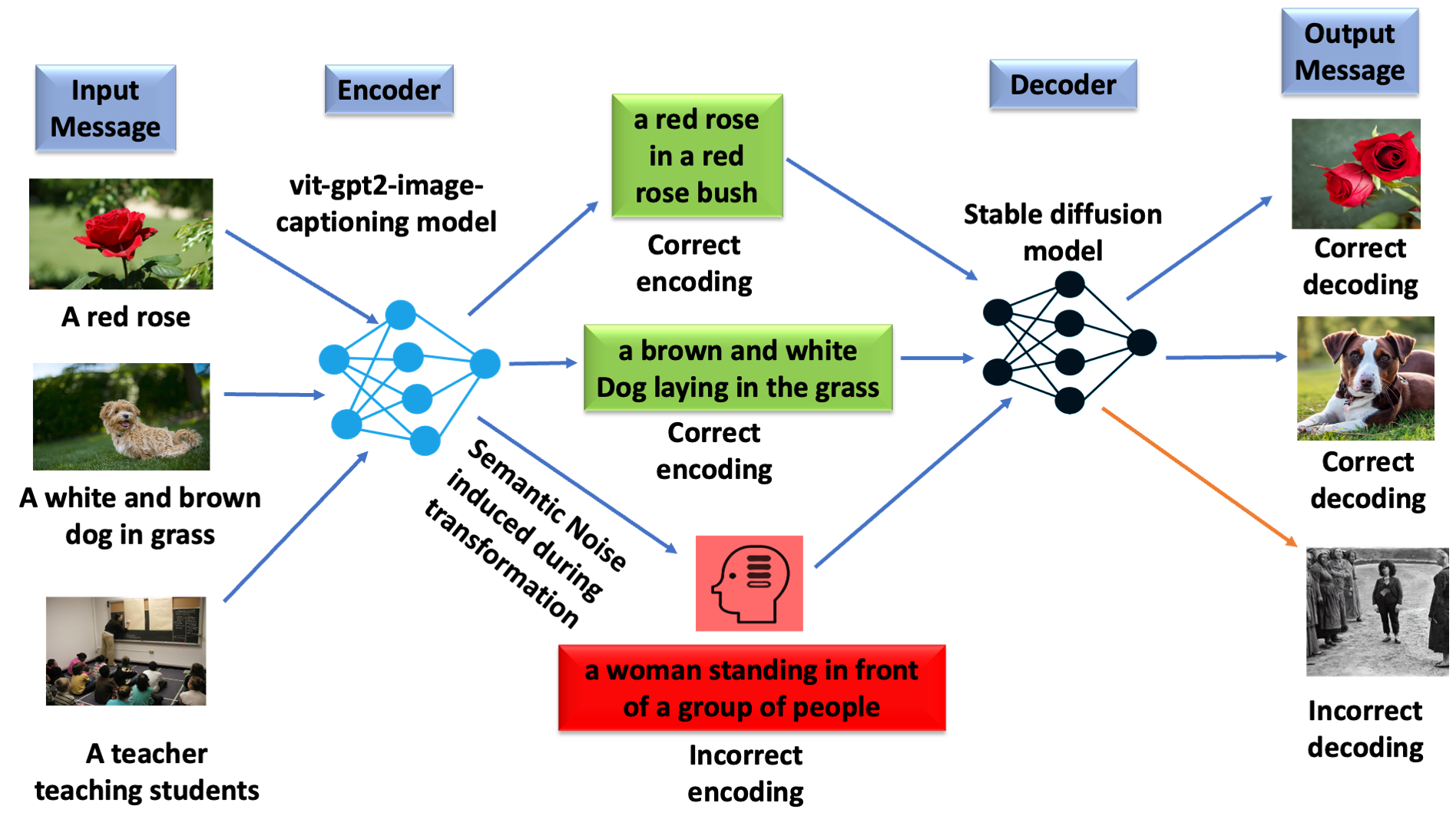}}
\caption{Effect of semantic noise during semantic transformation}
\label{fig:SemComm_EncDec}
\end{figure*}

\section{EOSL Adaptation to Diverse Semantic Tasks with Cumulative Learning}
\label{sec:cumulative_learning}

\subsection{Meta Learning}

Meta-learning, also known as ``learning to learn," is a subfield of machine learning that focuses on training models to quickly adapt to new tasks, datasets, or environments with minimal additional training data~\cite{lake2023human,verma2020meta}. Regular learning involves acquiring knowledge and skills from data. Meta-learning goes a step further by enabling the system to improve its learning process across different tasks. In the context of machine learning, this means that a model is said to have meta-learned if it can leverage its previous experiences and learning to improve its performance on new, unseen tasks or datasets. Meta-learning models aim to learn new tasks or datasets with only a few examples or iterations and are designed to be flexible and adaptable across a wide range of functions and domains. Key Characteristics Meta-learning includes:

\begin{itemize}
\item Few-shot learning: Meta-learning models aim to learn new tasks or datasets with only a few examples or iterations.
\item Task-agnostic: Meta-learning models are designed to be flexible and adaptable across a wide range of tasks and domains, including tasks with varying complexity, data distributions, label spaces, modalities, and objectives.
\item Learning to learn: Meta-learning models aim to improve their learning abilities, rather than simply memorizing new information.
\end{itemize}

For example, a meta-learning model trained on multiple image classification tasks can quickly adapt to a new task with only a few examples or a meta-learning model trained on multiple languages can promptly learn to translate a new language with minimal additional training data.

\subsection{Applying Meta Learning Principles to EOSL}
It is possible to demonstrate that utilizing EOSL in a cumulative manner enables the model selection system to accommodate the selection of an appropriate model even in diverse and varying contexts. This cumulative learning process draws inspiration from Meta-Learning~\cite{dahlhaus2006statistical,rajeswaran2019meta,wilamowski2010neural} but diverges from traditional approaches by leveraging historical EOSL values instead of active parameter optimization or backpropagation methods. The EOSL-based model selection system adapts to context generalization by guiding the learning process with previous rounds' historical EOSL values, allowing it to accommodate varying contexts, as shown later in this paper. Notably, the selected transformer models can be compared to the Student Model in Meta-Learning, while the EOSL-based model selection system resembles the Instructor or Teacher Model, guiding the selection process and adapting to new contexts.

Let \( e_0 \) be the initial EOSL for a specific topic, with no historical EOSL. In this explanation, \( e_i \) represents the EOSL at the \( i \)-th round based on its current value, where \( i \in \mathbb{N}^+ \), while \( e_i' \) denotes the cumulative EOSL, incorporating interactions with historical EOSL from previous rounds.

Initially, \( e_0 \) and \( e_0' \) are the same because there is no history.

\textbf{Initial EOSL (\( e_0 \)):}
\[
e'_0 = e_0
\]

In the later rounds, we introduce two different weight parameters, \(\alpha\) and \(\beta\). These are the weight coefficients associated with the current EOSL and historical EOSL, respectively. Here, EOSL is a positive real number and \(\alpha, \beta \in (0, 1) \setminus \{0, 1\}\) such that \(\alpha + \beta = 1\).

\textbf{EOSL at the 1st round:}
\begin{equation}
e'_1 = e_1\alpha + e_0\beta
\label{eq:e1_prime}
\end{equation}

\textbf{EOSL at the 2nd round:}
\begin{equation}
e_2' = e_2  \alpha + e_1'  \beta
\end{equation}

Which can be written as:
\begin{equation}
e_2' = e_2  \alpha + e_1  \alpha  \beta + e_0  \beta^2
\label{eq:e2_prime}
\end{equation}

\textbf{EOSL at the 3rd round:}
\begin{equation}
e_3' = e_3  \alpha + e_2'  \beta
\end{equation}

Which then again can be re-written as:
\begin{equation}
e_3' = e_3  \alpha + e_2  \alpha  \beta + e_1  \alpha  \beta^2 + e_0  \beta^3
\label{eq:e3_prime}
\end{equation}

By observing the equations (\ref{eq:e1_prime}), (\ref{eq:e2_prime}) and (\ref{eq:e3_prime}) we can write the $n^{th}$ term as:

\begin{equation}
e'_n = e_n  \alpha + e_{n-1}  \alpha  \beta + e_{n-2}  \alpha  \beta^2
+ \ldots + e_1  \alpha  \beta^{n-1} + e_0  \beta^n
\end{equation}

which then can be written as:

\begin{equation}
e'_n = \alpha  (e_n + e_{n-1}\beta + e_{n-2}  \beta^2 + \ldots \\
+ e_1  \beta^{n-1}) + e_0  \beta^n
\label{eq:en_prime}
\end{equation}

So the general expression for \( n \)-th EOSL is:

\begin{equation}
e_n' = \alpha \sum_{i=0}^{n-1} e_{n-i}\beta^i + e_0 \beta^n
\label{eq:en_general}
\end{equation}

It can be observed from (\ref{eq:en_prime}) or (\ref{eq:en_general}), as \( n \) becomes larger, the effect of older EOSL values diminishes due to the increasing powers of \( \beta \) in the expression. This means that the contribution of the EOSL from the current round and recent rounds becomes more significant compared to EOSL values from earlier rounds. This behavior is consistent with the weighting factors \( \alpha \) and \( \beta \) controlling the influence of current and historical EOSL values. Since \( \alpha \) and \( \beta \) are both fractions less than one, their effects gradually diminish with each additional round, making the EOSL calculation more reliant on recent data.

\section{Semantic Encoder and Decoder Design}
\label{sec:semantic_encoder_decoder}
We primarily designed the Encoder and Decoder system blocks and performed testing to exhibit and assess the semantic noise ($N_{sm}$).  We used pre-trained transformer-based model checkpoints hosted in the Hugging Face public repository~\cite{wolf2020transformers} to design our semantic encoder. Transformers have emerged as a prominent neural network architecture, specifically designed to tackle sequence-to-sequence tasks encompassing machine translation, text summarizing, and question answering. Leveraging an attention mechanism, transformers excel in capturing intricate relationships among diverse segments within a sequence. This characteristic makes them applicable not only to text-based scenarios but also enables their utilization in cross-modal tasks such as text-to-image and image-to-text conversions. In text-to-image and image-to-text conversion tasks, Transformers exhibit their capability to grasp long-range dependencies. By establishing associations between textual descriptions and corresponding image pixels, transformers acquire the capability to generate visually coherent images aligned with the provided textual input. Consequently, the inherent ability of transformers to facilitate inter-modality conversion renders them a highly capable choice for constructing the encoder and decoder components of a semantic communication system. 

This transformer model transforms an image into text, to be transmitted via a communication channel. We evaluated several encoders. On the other end, we used another neural network model, the Stable Diffusion Model~\cite{rombach2022high} to design our semantic decoder for text to image. We only used one semantic decoder; evaluations of decoders will be the focus of future work. We followed a CUDA-enabled implementation, initially developed using CLIP (Contrastive Language-Image Pre-Training) by openAI~\cite{radford2021learning} and piped that to the CPU.

\begin{figure*}[hbt!]
\centerline{\includegraphics[width=.9\textwidth]{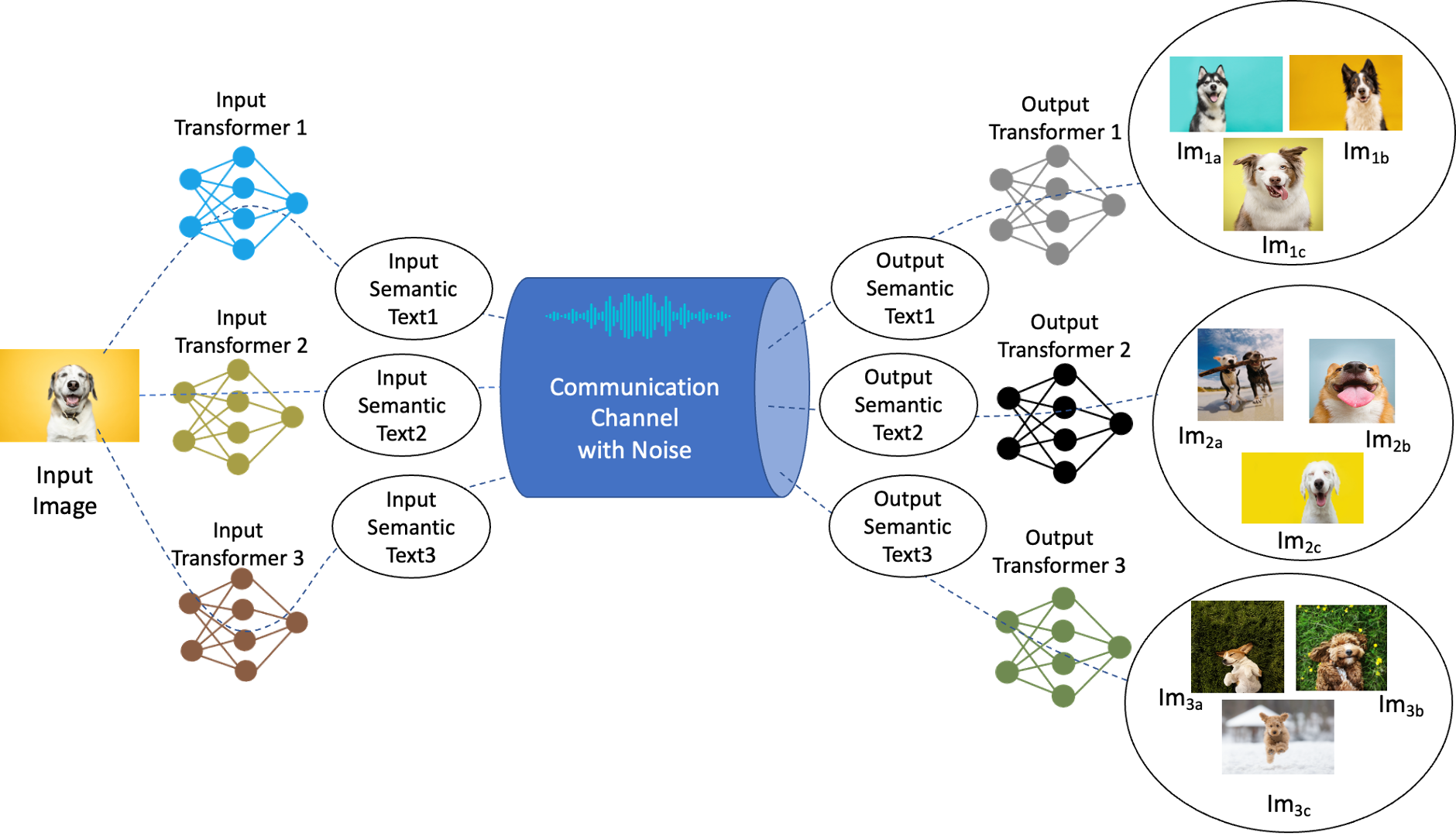}}
\caption{End to end image/text transformer based SemCom with communication channel}
\label{fig:select}
\end{figure*}

As illustrated in Fig.~\ref{fig:SemComm_EncDec}, we tested the semantic encoder and decoder with three images, and it successfully decoded the image correctly to the appropriate text for the first two messages of “A red rose” and “A white and brown dog in grass”. After decoding, the semantics were preserved from the first two messages, so when they were again decoded using our semantic decoder, going from text to image, they were able to preserve the semantics of the input message. But the third message, which is a picture of “A teacher teaching students”, was incorrectly encoded by our semantic encoder; this is an example of semantic or cognitive noise, which occurred due to misinterpretation by the encoder. As a result, when this text was again decoded by our semantic decoder it was transformed into an image having different semantics.

Fig.~\ref{fig:select}  illustrates a possible scenario for transformer model selection in semantic communication. A single input image can be encoded into text using multiple transformer options, and after transmission through a noisy communication channel, these text encodings may produce different semantic interpretations on the receiving end. Furthermore, when decoding and regenerating the text back into images, multiple transformer options (used as decoders) may be available, resulting in a multitude of possible output images for each semantic text. In this figure, only three encoder transformer and three decoder transformer options are shown, demonstrating how a single input image can yield nine distinct output images with potentially vastly different semantic meanings for the end user.

\begin{figure}[hbt!]
\centerline{\includegraphics[width=\columnwidth]{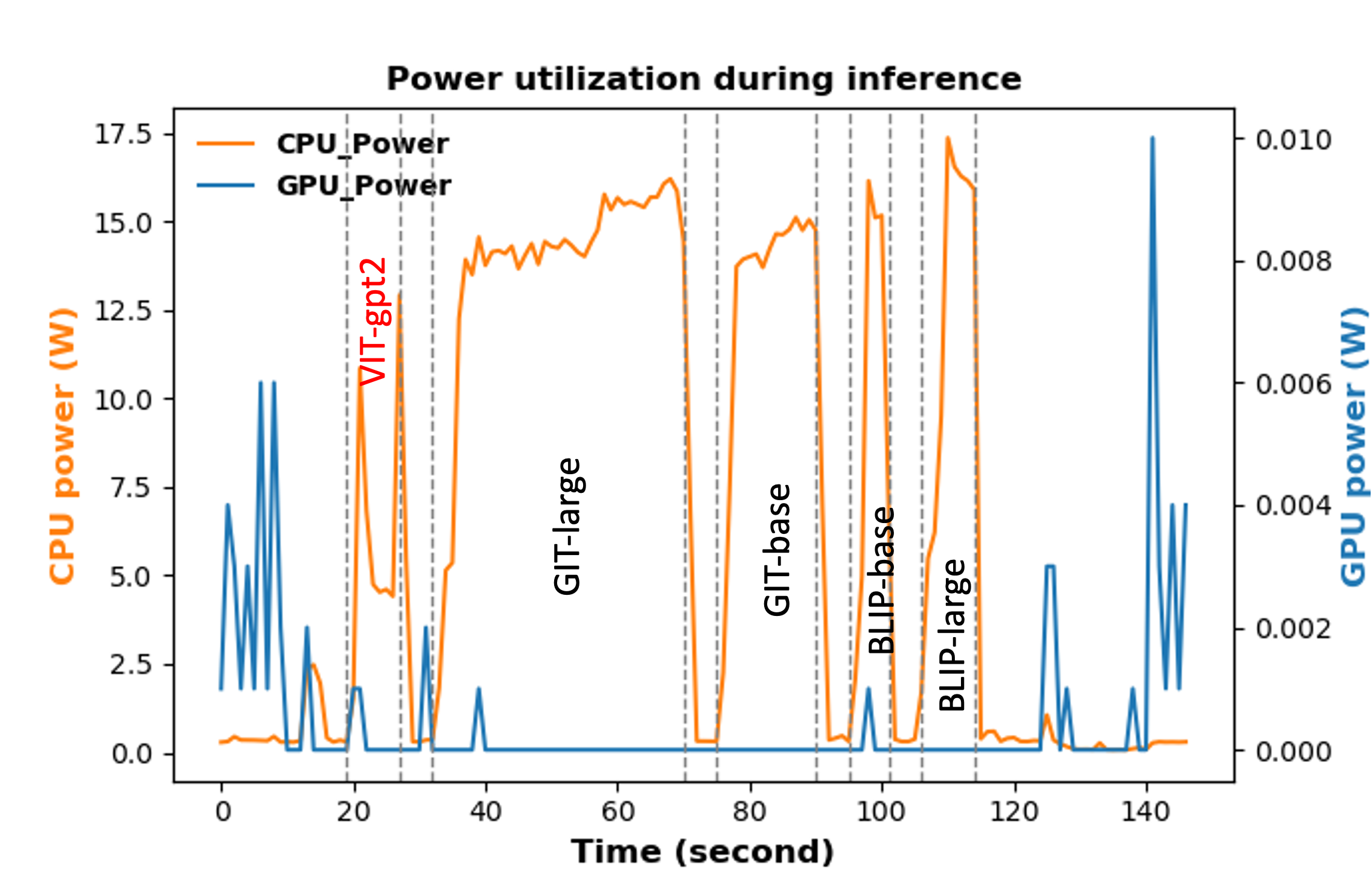}}
\caption{Resource Utilization During Inference}
\label{fig:tfm_comp}
\end{figure}

\section{Experimental Results}
\label{sec:results}
In this section, we present two main sets of results: (1) an encoder-based experiment and (2) a combined encoder-decoder experiment. Additionally, we expand our investigation to assess the generalization capabilities of EOSL on a diverse and comprehensive image dataset in part (3).

\subsection{Image-to-text encoding and EOSL-based model selection}
We have chosen five different caption generator transformer models to perform detailed energy benchmarking experiments during the image-to-text generation inference task. Specifically, we utilized the five encoder models `BLIP-base (Bootstrapping Language-Image Pre-training),' `GIT-base (Generative Image Transformer),' `GIT-large,' `BLIP-large,' and `VIT-GPT-2 (Vision Transformer),' to convert the image into text. These models were run locally on an Apple M1 chipset MacBook Air with 8~GB memory and 256~GB storage using the MacOS Ventura operating system. It had a total of 8 cores (4 performance and 4 efficiency). We used a MacOS-based CLI utility `Powermetrics' to collect raw energy utilization data while performing individual model inferences.

\begin{figure*}[hbt!]
\centerline{\includegraphics[width=\textwidth]{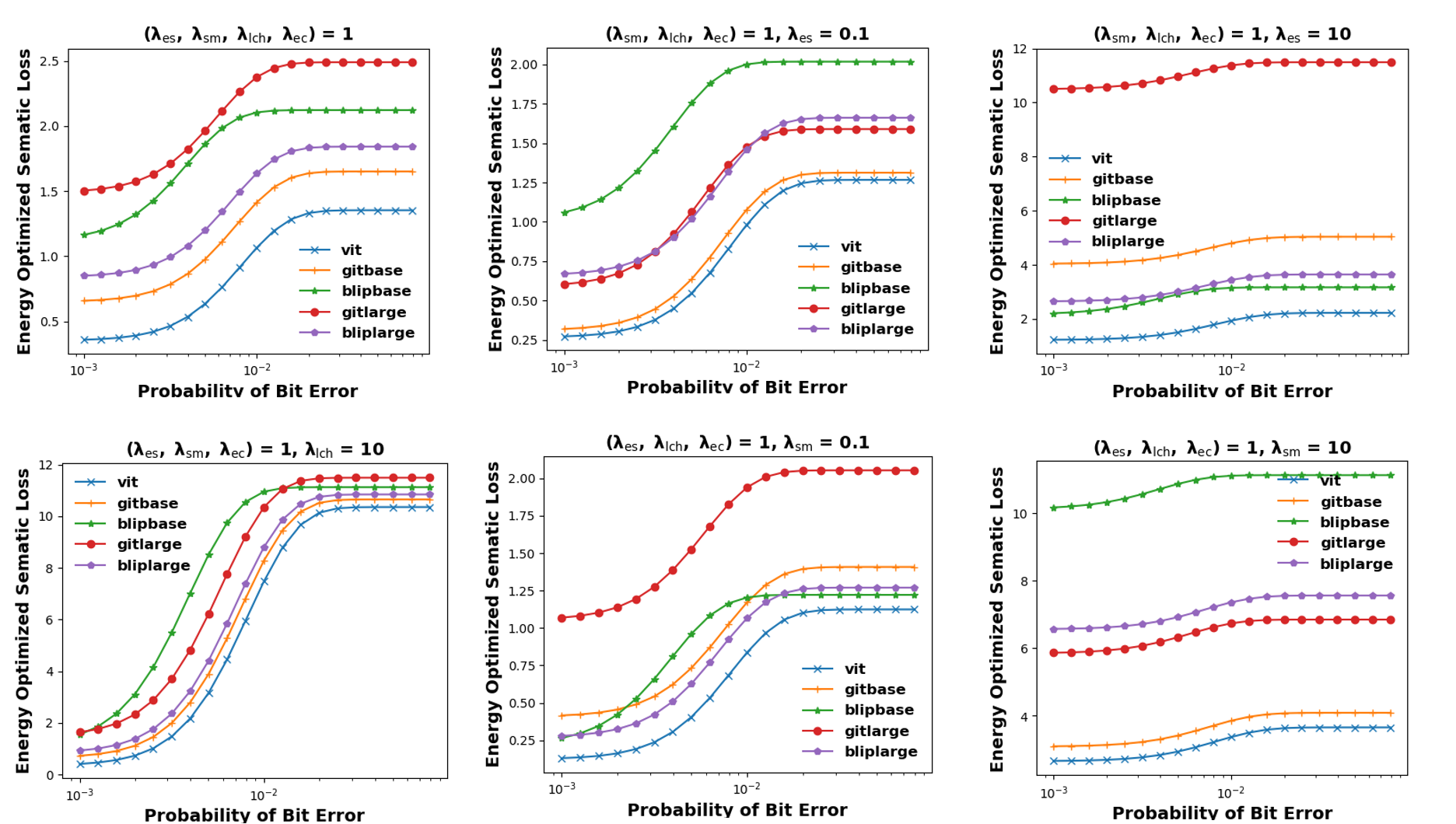}}
\caption{Changes of EOSL with the probability of bit error rate when using different values of $\lambda_{sm}$, $\lambda_{lch}$, $\lambda_{e_c}$, $\lambda_{e_s}$}
\label{fig:eosl_all}
\end{figure*}

We selected a high-resolution (14 Mb) image of a dog as an input for the caption generation task (seen in Fig.~\ref{fig:dogs}) and used the 5 models to generate 5 different captions from the same image. We also defined a text description of the image `a brown dog running through grassy field', used as the correct semantics of this image. Those 5 generated captions were compared for text-based cosine similarity with our defined semantics. This gave us five different semantic similarity scores, from which we could calculate the semantic noise using equation~(\ref{eq:Nsm}). We also recorded CPU and GPU utilization performance data alongside the timestamps and duration for each model inference. All the consumption data was collected in 1-second intervals. Finally, we accumulated the most relevant parameters like total CPU and GPU energy (in Joules), CPU utilization~\%, etc., then plotted them with respect to time in seconds on the $x$-axis as shown in Fig.~\ref{fig:tfm_comp}. The stop and start of each model inference for various transformers has been shown using grey-dotted vertical lines in Fig.~\ref{fig:tfm_comp}.  As observed, larger models like GIT-large or BLIP-large had much higher energy footprints, but base models like VIT-GPT-2 or GIT-base consumed much less resources in terms of power and CPU utilization. The total energy consumed during an inference was obtained from the summation of instantaneous power as below:
\begin{equation}
E = \sum_{j=1}^{n} \sum_{i=1}^{m} P_{ij}\Delta t = \sum_{j=1}^{n} \sum_{i=1}^{m} P_{ij}
\end{equation}

\noindent Values are reported for $\Delta t=1$ sec, and $P_{ij}$ is the instantaneous power at $i^{th}$ second during $j^{th}$ transmission. 

When EOSL is plotted against communication bit error probability for various transformer models in Fig.~\ref{fig:eosl_all}, it can be observed that the VIT-GPT-2 model (``vit") maintained the lowest EOSL with the increase in bit error probability for every scenario. Moreover, given that these models primarily operate on CPUs, the utilization of GPU power is minimal in comparison to that of the CPU, as depicted in Fig.~\ref{fig:tfm_comp} and Table~\ref{tab1}. There we assumed a fixed average bit error probability of $0.001$, a data rate of 143 Mbps (the rate per 20~MHz in IEEE 802.11ax), maximum admissible power of 1 Watt as regulated by FCC 15.247, the average packet size of 1500 bytes as in a traditional communication system, and all the weight parameters are set to $\lambda$=1. Based on the results shown in Table~\ref{tab1}, VIT-GPT2, and GIT-base had lowest semantic noise; both are below $N_{sm_{\text{thresh}}}=0.3$. However, we assume in the experiment $N_{sm_j} \leq N_{sm_{\text{thresh}}}$ is satisfied at $i=1$ for all cases; no re-transmission was involved.

\begin{table*}[htbp] 
\caption{Energy consumption during inference and EOSL values}
\centering
\footnotesize 
\setlength{\tabcolsep}{3pt} 
\begin{tabular}{|c|c|c|c|c|c|}
\hline
\textbf{Encoder} & \textbf{Total CPU Energy (J)} & \textbf{Total GPU Energy (J)} & \textbf{Total CPU Utilization (\%)} & \textbf{Semantic Noise} & \textbf{EOSL} \\
\hline
\textbf{VIT-GPT2} & \textbf{50.701} & \textbf{0.002} & \textbf{571.9} & \textbf{0.255} & \textbf{0.360} \\
\hline
BLIP-base & 60.922 & 0.001 & 513.4 & 1.000 & 1.164 \\
\hline
GIT-base & 197.442 & 0 & 1456.1 & 0.270 & 1.504 \\
\hline
BLIP-large & 105.095 & 0 & 746.3 & 0.635 & 0.659 \\
\hline
GIT-large & 524.718 & 0.001 & 3669.9 & 0.484 & 0.850 \\
\hline
\end{tabular}
\label{tab1}
\end{table*}

\subsection{Image-to-Text Encoding and Text-to-Image Decoding}

\begin{table*}[htbp]
    \caption{Encoder Size and Complexity with Semantic Efficiency}
    \centering
    \small
    \setlength{\tabcolsep}{2pt}
    \begin{tabular}{|c|c|c|c|c|c|c|c|}
        \hline
        \textbf{\textit{Encoder}} & \textbf{\textit{Decoder}} & \textbf{\textit{Size (Mb)}} & \textbf{\textit{Parameters (M)}} & \textbf{\textit{Cosine Similarity}} & \textbf{\textit{SSIM}} & \textbf{\textit{EOSL (cosine)}} & \textbf{\textit{EOSL (SSIM)}} \\
        \hline
        GIT-base & \multirow{5}{*}{Stable-Diffusion} & 673 & 177 & 0.878 & 0.654 & 0.321 & 0.123 \\
        \cline{1-1} \cline{3-8}
        \textbf{VIT-GPT2} & & \textbf{936} & \textbf{239} & \textbf{0.842} & \textbf{0.556} & \textbf{0.189} & \textbf{0.234} \\
        \cline{1-1} \cline{3-8}
        BLIP-base & & 943 & 247 & 0.837 & 0.530 & 0.267 & 0.321 \\
        \cline{1-1} \cline{3-8}
        GIT-large & & 1503 & 394 & 0.836 & 0.592 & 0.412 & 0.543 \\
        \cline{1-1} \cline{3-8}
        BLIP-large & & 1791 & 470 & 0.822 & 0.238 & 0.524 & 0.345 \\
        \hline
    \end{tabular}
    \label{tab2}
\end{table*}

\begin{figure*}[hbt!]
\centerline{\includegraphics[width=\textwidth]{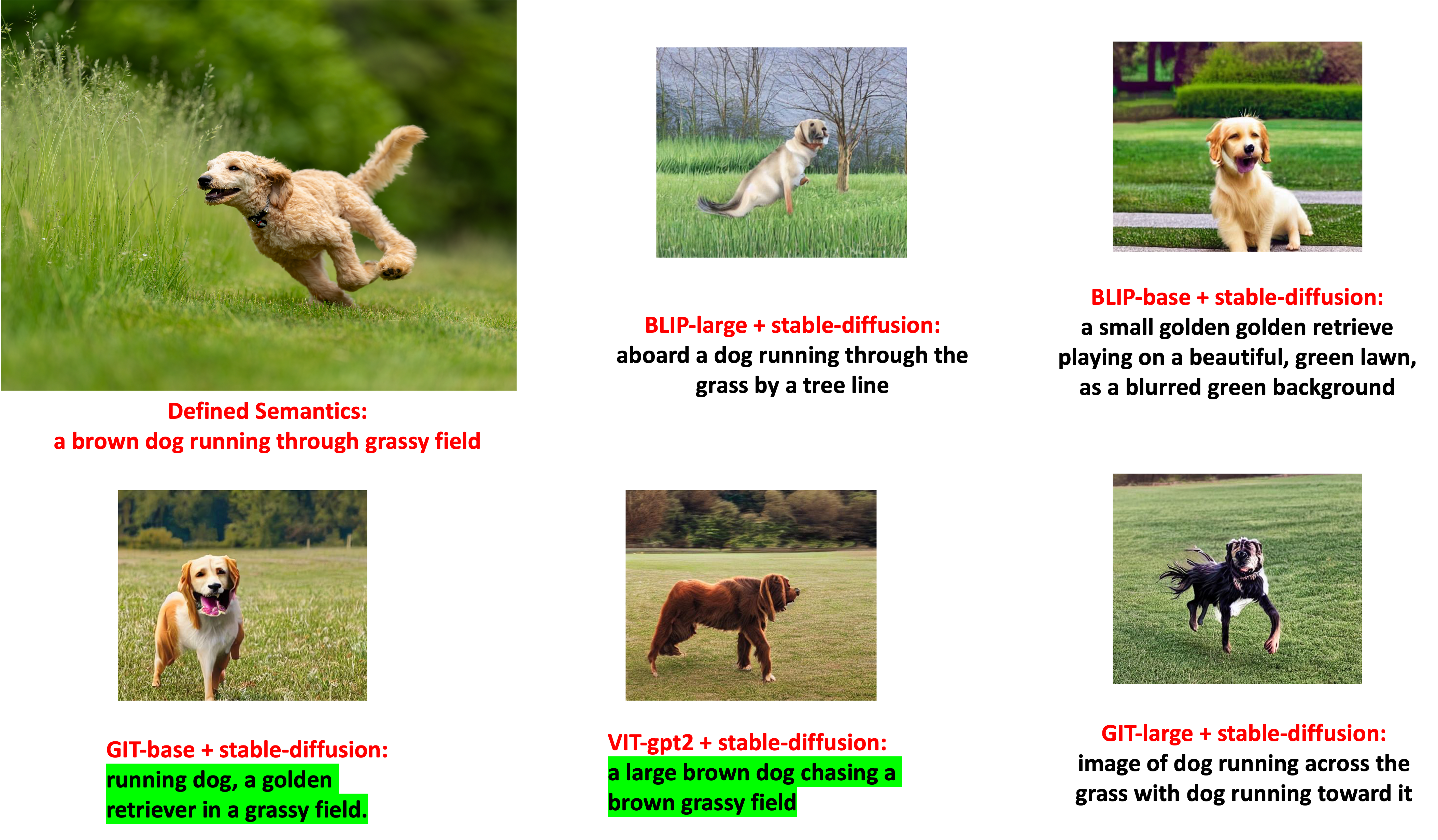}}
\caption{Illustration of semantic efficiency of different encoder models and images generated by diffusion}
\label{fig:dogs}
\end{figure*}

We conducted another experiment involving the transformation of a sample image to text and then from text to image using transformer models for both input and output stages. The same five encoder models, as mentioned in the first experiment, were used to convert the image into text. Additionally, in this case, a single decoder model, which is a text-to-image generator transformer named `Small-Stable-Diffusion-v0' was deployed for the reverse transformation, i.e. text to image generation. Fig.~\ref{fig:dogs} shows the main image, the semantics below it, and the text generated by all the five models are shown, along with the images generated from each text by stable diffusion model. Interestingly, our findings from Table~\ref{tab2} reveals that the similarity metrics are not dependent on or influenced by the sizes of the models utilized. This observation held when we repeated the experiment with images having their background removed, obtaining similar results. Remarkably, based on the outcomes of our current experiments, the `VIT-GPT2' encoder model emerged as the most promising candidate, as it exhibited superior semantic efficiency across all types of similarity comparisons presented in Table~\ref{tab2}. This finding is notable considering that the `VIT-GPT2' model is relatively smaller in size and possesses fewer hyper-parameters compared to larger and more complex alternatives such as `BLIP-Large' and `GIT-Large,' as shown in Table~\ref{tab2}. It should be noted that the semantic decoder model consumed approximately 40 times more energy than the 5 encoder models.  Text-to-image creation consumed an average of approximately 4~kJ which would heat 10 ml of water from room temperature to boiling. Further evaluation of text-to-image creation is a subject of future work.

\subsection{Evaluating EOSL Performance with Context Variation}

\begin{figure*}[hbt!]
\centerline{\includegraphics[width=5 in]{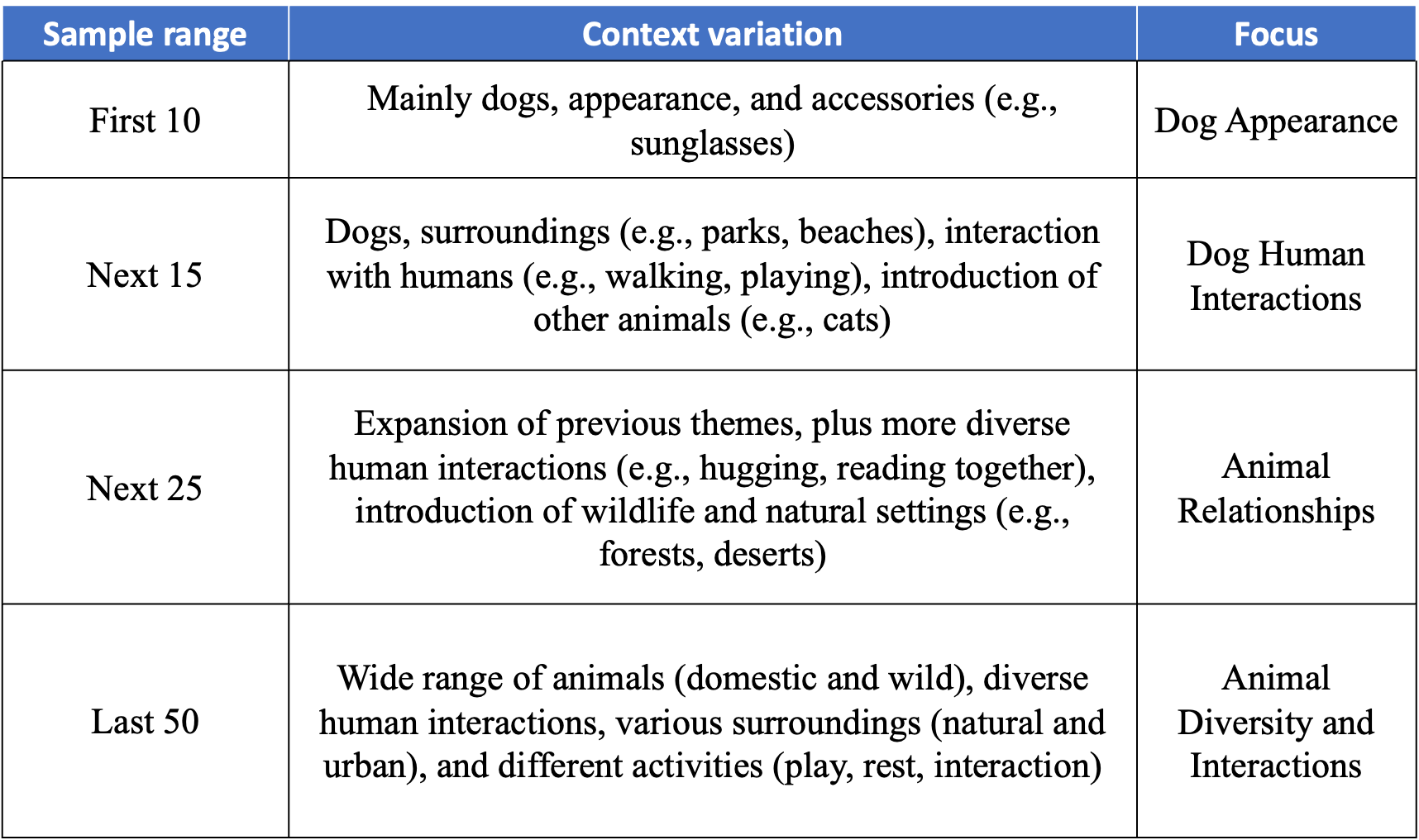}}
\caption{Variation of topic context}
\label{fig:context}
\end{figure*}

In this experiment, we test EOSL's capability to select the right model even with cumulatively varying subjects. To test this, we collected a few images of a subject, then cumulatively added more and more images of more diverse contexts or topics. As shown in the table in Fig.~\ref{fig:context}, the first 10 images are of a single subject, mainly focusing on the appearance of a dog. The next 15 images increase topic diversity by adding humans and their interactions with dogs. The next 25 images further diversify the topic by adding more diverse interactions and surrounding details, along with inter-animal relationships by adding more animal images to the data. Finally, we added 50 new images that are of various animals but not dogs.

\begin{figure*}[hbt!]
\centerline{\includegraphics[width=\textwidth]{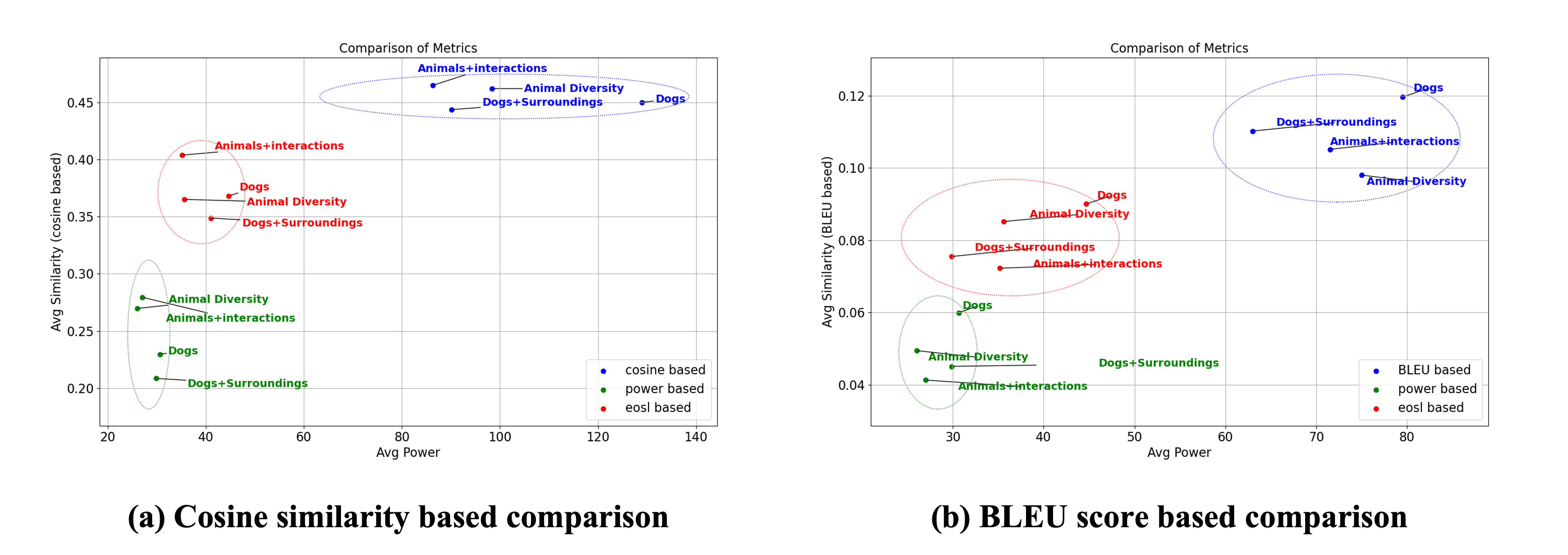}}
\caption{Avg. Similarity vs. Avg. Power Performance Comparison of Metrics}
\label{fig:mets}
\end{figure*}

\begin{table*}[htbp]
    \centering
    \caption{EOSL-Based Leader Board at Initial Step (First 10 samples)}
    \label{tab:leader}
    \begin{tabular}{cccccc}
        \toprule
        Image & Winner Model & EOSL & Total Power Spent (J) & Cosine Similarity \\
        \midrule
        1.jpg & blip-image-captioning-base & 0.7627 & 41.5830 & 0.4082 \\
        2.jpg & blip-image-captioning-large & 1.1470 & 65.8970 & 0.5071 \\
        3.jpg & blip-image-captioning-base & 0.6658 & 32.2100 & 0.4629 \\
        4.jpg & blip-image-captioning-large & 1.1316 & 58.9600 & 0.1581 \\
        5.jpg & blip-image-captioning-base & 0.7168 & 30.1130 & 0.3780 \\
        6.jpg & vit-gpt2-image-captioning & 0.8568 & 47.7020 & 0.4000 \\
        7.jpg & git-base-coco & 0.8960 & 42.7010 & 0.2357 \\
        8.jpg & blip-image-captioning-base & 0.8975 & 21.0200 & 0.2500 \\
        9.jpg & git-base-coco & 0.8795 & 75.6270 & 0.3780 \\
        10.jpg & blip-image-captioning-base & 0.7016 & 30.9340 & 0.5040 \\
        \bottomrule
    \end{tabular}
\end{table*}

\begin{table*}[htbp]
\caption{Similarity to Power Ratio (SPR) for Different Metrics Using Cosine Similarity}
\centering
\small
\setlength{\tabcolsep}{1pt}
\begin{tabular}{|c|c|c|c|}
\hline
\textbf{Sample Size} & \textbf{Similarity Based SPR} & \textbf{Power Based SPR} & \textbf{EOSL Based SPR} \\
\hline
10 & $3.4926 \times 10^{-3}$ & $7.4946 \times 10^{-3}$ & $8.2417 \times 10^{-3}$ \\
25 & $4.9258 \times 10^{-3}$ & $6.9847 \times 10^{-3}$ & $8.5124 \times 10^{-3}$ \\
50 & $5.3902 \times 10^{-3}$ & $10.3514 \times 10^{-3}$ & $11.4882 \times 10^{-3}$ \\
100 & $4.6983 \times 10^{-3}$ & $10.3662 \times 10^{-3}$ & $10.8188 \times 10^{-3}$ \\
\hline
\end{tabular}
\label{cosine}
\end{table*}

\begin{table*}[htbp]
\caption{Similarity to Power Ratio (SPR) for Different Metrics Using BLEU Score}
\centering
\small
\setlength{\tabcolsep}{1pt}
\begin{tabular}{|c|c|c|c|}
\hline
\textbf{Sample Size} & \textbf{Similarity Based SPR} & \textbf{Power Based SPR} & \textbf{EOSL Based SPR} \\
\hline
10 & $1.5050 \times 10^{-3}$ & $1.9564 \times 10^{-3}$ & $2.0154 \times 10^{-3}$ \\
25 & $1.7493 \times 10^{-3}$ & $1.5118 \times 10^{-3}$ & $2.5305 \times 10^{-3}$ \\
50 & $1.4700 \times 10^{-3}$ & $1.5298 \times 10^{-3}$ & $2.0551 \times 10^{-3}$ \\
100 & $1.3072 \times 10^{-3}$ & $1.9014 \times 10^{-3}$ & $2.3925 \times 10^{-3}$ \\
\hline
\end{tabular}
\label{bleu}
\end{table*}

Now, we will use different metrics to select the best-performing model for each image of a set and declare a winner per image from the five image-to-text transformer models we have been using. The task of the transformer here is to generate a caption from a given image. We compare the similarity performance of a model by comparing its generated caption to pre-defined captions. These pre-defined captions are previously user-chosen as per their suitable description from various caption generators and alt-text generator websites. For each caption generation task, we measure several parameters such as similarity score (using cosine similarity as well as BLEU score), and we measure the CPU and GPU power in Watts used for that transformation task by that particular model. We also compute EOSL. We repeat this process of collecting these parameters in each round for 10, 25, 50, and 100 images with increasingly diverse contexts.

Now, in each round, based on the above calculations, we choose various winner leader boards. These leader boards have the best-performing model based on minimum power usage for each image, based on maximum similarity or BLEU score achieved, and based on the lowest EOSL value. One such example of leader board based on lowest EOSL value has been shown in Table~\ref{tab:leader}. The purpose of these leader boards is to compare how each metric performs in selecting the best models for each image-to-text conversion. After the base round with 10 images, every subsequent round of 25, 50, and 100 EOSL values are governed by the previous rounds using equation (\ref{eq:en_general}). For this experiment, the value of \( \alpha \) was set to 0.7 and the value of \( \beta \) was set to 0.3

Our analysis of the leader boards revealed that models selected based on minimum power usage had low power consumption but poor similarity performance, while models chosen based on similarity metrics (cosine similarity or BLEU scores) achieved higher average similarity values but consumed more power. Notably, EOSL successfully identified models with high similarity scores while maintaining minimal power usage. This can be observed using an average similarity to average power ratio (SPR); SPR is calculated as the ratio of average similarity to average power consumption across all samples in a leaderboard. EOSL outperformed other comparison metrics in terms of SPR. Specifically, EOSL achieved upto 136\% better SPR compared to cosine similarity only metrics and 22\% better SPR compared to minimum power based metrics (Table~\ref{cosine}), and 83\% better SPR compared to BLEU score only metrics and 67\% better SPR compared to minimum power based metrics (Table~\ref{bleu}). This can also be observed in Fig.~\ref{fig:mets} which shows the relationship between average power consumption average similarity at each round of EOSL calculation.

Moreover, EOSL's performance remained robust even with changes in context across rounds, demonstrating its superiority and adaptability compared to other approaches.

\section{Conclusion and Future Work}
\label{sec:conclusion}
In conclusion, our research demonstrates the promising potential of Energy-Optimized Semantic Loss (EOSL) in semantic communication, opening up new avenues for innovation in this field. By introducing an innovative multi-objective loss function, we harmoniously balance semantic information loss and energy consumption. Our comprehensive experiments demonstrate that EOSL-based encoder model selection achieves notable semantic efficiency without excessive computational and energy resources. Notably, our approach diverges from the trend of increasingly complex tasks requiring more complex models, instead enabling intricate tasks like semantic transformations with superior semantic efficiency while ensuring limited energy consumption. Inspired by Meta-Learning principles, we successfully extend the applicability of EOSL to diverse and varying contexts without requiring additional backpropagation, a useful contribution to the field. Our results show that EOSL consistently outperforms other metrics in terms of similarity-to-power ratio, even with changes in context.

It is also possible to further enhance the generalization capabilities of adaptation techniques like Retrieval-augmented generation (RAG) and Low-Rank Adaptation (LoRA) by applying the concept of continual learning explored by our current research, where the model retains knowledge from previous tasks. By leveraging historical knowledge and incrementally adapting to new tasks and contexts, RAG and LoRA can improve their efficiency and effectiveness in few-shot learning scenarios, leading to more robust and adaptable AI systems. This combined approach has the potential to enhance the sustainability of communication systems, paving the way for a new generation of effective and environmentally friendly solutions. While there is still room for future improvements, such as exploring a broader range of semantic datasets and fine-tuning transformer parameters for optimal EOSL and energy trade-offs, our work lays a solid foundation for energy-efficient neural network selection and the development of green semantic communication architectures.

\bibliography{bibtex/bib/reference}
\bibliographystyle{IEEEtran}

\end{document}